\pdfoutput=1

\documentclass[11pt]{article}

\usepackage{EMNLP2022}

\usepackage{times}
\usepackage{latexsym}

\usepackage[T1]{fontenc}

\usepackage[utf8]{inputenc}

\usepackage{inconsolata}

\usepackage{url}

\usepackage{caption}
\usepackage{amsmath}
\usepackage[ruled,linesnumbered]{algorithm2e} 
\usepackage{amssymb}
\usepackage{bm}
\usepackage{empheq}
\usepackage{adjustbox}
\usepackage{tabulary} 
\usepackage{multirow}
\usepackage{makecell}
\usepackage{color} 
\usepackage{setspace} 

\usepackage{xcolor}
\definecolor{green_}{HTML}{DDF4EA}
\definecolor{blue_}{HTML}{DEEFF9}
\definecolor{pink_}{HTML}{FFF0DD}
\definecolor{grey_}{HTML}{C4CBCF}
\usepackage[normalem]{ulem} 
\newcommand\hlgreen{\bgroup\markoverwith
  {\textcolor{green_}{\rule[-.5ex]{2pt}{2.5ex}}}\ULon}
\newcommand\hlblue{\bgroup\markoverwith
  {\textcolor{blue_}{\rule[-.5ex]{2pt}{2.5ex}}}\ULon}
\newcommand\hlpink{\bgroup\markoverwith
  {\textcolor{pink_}{\rule[-.5ex]{2pt}{2.5ex}}}\ULon}
\newcommand\hlgrey{\bgroup\markoverwith
  {\textcolor{grey_}{\rule[-.5ex]{2pt}{2.5ex}}}\ULon}

\usepackage{pbox}
\usepackage{graphicx, subfigure}
\usepackage{booktabs}
\usepackage{tikz}
\usetikzlibrary{fit,positioning}
\usetikzlibrary{arrows}
\usepackage{etoolbox}
\newcommand{\circled}[2][]{\tikz[baseline=(char.base)]
    {\node[shape = circle, draw, inner sep = 0.0pt]
    (char) {\phantom{\ifblank{#1}{#2}{#1}}};%
    \node at (char.center) {\makebox[0pt][c]{#2}};}}
\robustify{\circled}
\newcommand{\tabincell}[2]{\begin{tabular}{@{}#1@{}}#2\end{tabular}}  

\def\aclpaperid{0000} 

\title{
  Semantic-aware Contrastive Learning for 
   More Accurate Semantic Parsing
}

\begin{document}

\author{
  Shan Wu$^{1,2}$,
  Chunlei Xin$^{1,2}$,  
  Bo Chen$^{1}$,
  Xianpei Han$^{1,3,}$\thanks{~ Corresponding Author} $^{\,}$,
  Le Sun$^{1,3}$ $^{\,}$  \\
  $^{1}$Chinese Information Processing Laboratory ~ 
  $^{2}$University of Chinese \\ 
  Academy  of Sciences, Beijing, China 
  $^{3}$State Key Laboratory of Computer Science \\
   Institute of Software, Chinese Academy of Sciences, Beijing, China \\
  \tt{\{wushan2018,chunlei2021,chenbo,xianpei,sunle\}@iscas.ac.cn} }

\maketitle

\begin{abstract}

Since the meaning representations are detailed and accurate annotations which express fine-grained sequence-level semtantics,
it is usually hard to train discriminative semantic parsers via Maximum Likelihood Estimation (MLE) in an autoregressive fashion. 
In this paper, we propose a semantic-aware contrastive learning algorithm, which can learn to distinguish fine-grained meaning representations and take the overall sequence-level semantic into consideration. 
Specifically, a multi-level online sampling algorithm is proposed to sample confusing and diverse instances. 
Three semantic-aware similarity functions are designed to accurately measure the distance between meaning representations as a whole. 
And a ranked contrastive loss is proposed to pull the representations of the semantic-identical instances together and push negative instances away. 
Experiments on two standard datasets show that our approach achieves significant improvements over MLE baselines and gets state-of-the-art performances by simply applying semantic-aware contrastive learning on a vanilla \textsc{Seq2Seq} model.

\end{abstract}

\section{Introduction}

Semantic parsing aims to translate natural language utterances into formal meaning representations(MRs), which has attracted much attention for many years \citep{Wong:2007,Kate:2005,WeiLu:2008,Guo:2019}. 
Recent studies mostly treat semantic parsing as a neural sequence to sequence translation task via encoder-decoder frameworks \citep{Dong:2016,Robin:2016,Rabin:2017,Chen:2018,Zhao:2020,Shao:2020}.

To train neural semantic parsers, most studies employ Maximum Likelihood Estimation, which optimizes the probabilities of the tokens in an autoregressive fashion. 
By decomposing sequence tasks into tokens, MLE training is good at n-gram based generation tasks such as machine translation and paraphrasing, but overlooks the semantics in a whole level.

\begin{figure}[t]
  \centering
  \noindent\makebox[.45\textwidth][c] {
  \includegraphics[width=.45\textwidth]{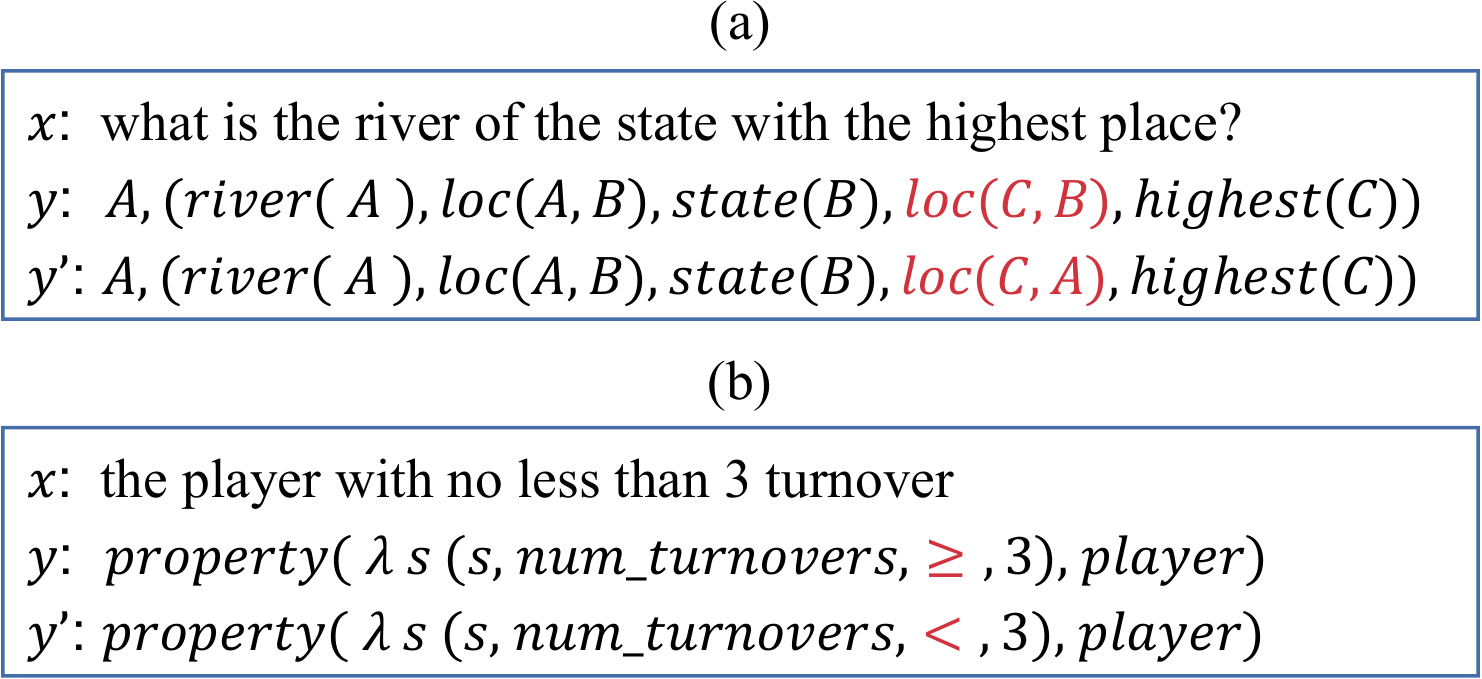}
  }
    \setlength{\belowcaptionskip}{-0.3cm}
  \caption{
 Meaning representations are detailed and accurate annotations which express fine-grained sequence-level semtantics. (a) the correct MR $y$ and the predicted $y'$ have similar token sequences but with very different semantics; (b) one token error (from $\ge$ to $<$) will reverse the semantics of MR.
  }
  \label{headFig-Examples}%
\end{figure}

Unfortunately, meaning representation is formal and detailed annotation, that is, it should be viewed as a sequence-level whole possessing fine-grained semantics. 
Such features make it hard to train accurate semantic parsers via MLE, which only computes loss token-by-token and is insensitive to small perturbations. 
For example,  in Figure \ref{headFig-Examples}(b) even one token error (from $\ge$ to $<$) can reverse the semantics of a meaning representation:  
\textit{Property ( $\lambda$ s  (s  num\_turnovers $\ge$ 3),  player)} to 
\textit{Property ( $\lambda$ s  (s  num\_turnovers $<$ 3),  player)}. 
For a case of pp-attatchment problem, the MRs in Fig  \ref{headFig-Examples}(a) have very similar token sequence but very different semantics.  
We analyze the error cases of a classical \textsc{Seq2Seq} parser in \textsc{Overnight} dataset$\footnote{More statistical details can be found in Appendix A.1.}$, and found that 
the edit distances  from 42.7\% of the error parses to the correct MRs are only 1, and that of 74.2\% of the error parses are  $\le$ 3. 
That is, most errors are due to the lack of ability to distinguish fine-grained semantics. 
Therefore it is crucial to develop learning algorithms which can take the sequence semantics and the fine-granularity of meaning representations into consideration.




In this paper, we propose Semantic-aware Contrastive Learning (SemCL), which can learn semantic-aware, fine-grained meaning representations for accurate semantic parsing. 
To resolve the fine-granularity challenge, we sample negative instances in different divergence levels. 
And a multi-level online sampling algorithm is proposed to collect confusing and diverse instances.  
To resolve the sequence-level semantics challenge, we compare meaning representations as a whole, rather in token-by-token. 
Three semantic-aware similarity functions are designed to accurately measure the distance between utterances and meaning representations. 
Finally, we propose \textit{ranked contrastive loss}, which is used to pull the representations of the semantic-identical instances together and push negative instances away (even if they look very similar to the positive ones). 
In this way, the semantic parsers can learn to distinguish fine-grained semantics and take the overall semantics into consideration.

In summary, the main contributions of this paper are:

\begin{itemize}
 \item We propose a semantic-aware contrastive learning algorithm, which can effectively model the fine-grained and sequence-level semantics in semantic parsing. To our best knowledge, this is the first attempt to adopt contrastive learning for semantic parsing.
 \item We design an effective contrastive learning algorithm, which contains a multi-level online sampling algorithm, three semantic-aware similarity functions, and a ranked contrastive loss. This framework can also benefit other tasks which depend on the distinguishing ability of the fine-grained or whole-level semantics.
 \item Experiments on two standard datasets show that our approach achieves significant improvements over MLE baselines, and gets state-of-the-art performances.
\end{itemize}

\section{Base \textsc{Seq2Seq} Parser}


This paper uses the classical \textsc{Seq2Seq} semantic parser as our base model due to its simplicity and effectiveness
\citep{Dong:2016}.

\paragraph{Encoder.}
Given a sentence 
$\mathbf{x} = w_1, w_2, ..., w_n$,
a bidirectional LSTM \citep{Hochreiter:1997} or BERT \citep{Devlin:2019} can be used to map words into 
$\mathbf{h}_\mathbf{x} = \mathbf{h}_1, \mathbf{h}_2, ..., \mathbf{h}_n$.

\paragraph{Attention-based Decoder.}

Given the sentence representation, the tokens of  the logical forms are generated sequentially. 
Specifically, the decoder is first initialized with the hidden states of the encoder. 
Then at each step $t$, let $\phi{(y_{t-1})}$ be the vector of the previous predicted token, the current hidden state $\mathbf{s}_{t}$ is obtained from $\phi{(y_{t-1})}$ and $\mathbf{s}_{t-1}$. 
We calculate the attentioned  source context representations for the current step $t$:
\begin{gather}
\alpha_t^i   = \frac{\exp{(\mathbf{s}_t\cdot\mathbf{h}_i})}{\sum_{i=1}^{n}\exp{(\mathbf{s}_t\cdot\mathbf{h}_i)}} \\
  \mathbf{c}_t  = \sum_{i=1}^n \alpha_t^i \mathbf{h}_i  
\label{ct}
\end{gather}
and the next token is generated from the vocabulary distribution:
\begin{gather}
  P(y_t|y_{<t},\mathbf{x}) =  \text{softmax}(\mathbf{W}_o[\mathbf{s}_t;\mathbf{c}_t]+\mathbf{b}_o)
\end{gather}




\paragraph{MLE Learning.}

The \textsc{Seq2Seq} model is trained by maximizing the likelihoods of  the tokens in an autoregressive fashion:
\begin{equation}
  \mathcal{L}_{seq2seq} = - \sum_{(\mathbf{x},\mathbf{y}) \in \mathbf{D}} 
  \sum_{t=1}^{m} \log p(y_t|y_{<t},\mathbf{x})
\end{equation}
where $\mathbf{D}$ is the corpus, $\mathbf{x}$ is the sentence, $\mathbf{y}$ is its logical form label.

However, such a token-by-token autoregressive training paradigm is insensitive to the overall semantics of the structured MR, making it hard to train effective and discriminative semantic parsers. 
We propose semantic-aware contrastive learning to help the semantic parsers to perceive the divergence of fine-grained semantics, which is overlooked in the existing autoregressive training approaches.
\begin{figure*}[!t]
  \centering
      \includegraphics[width=0.90\textwidth]{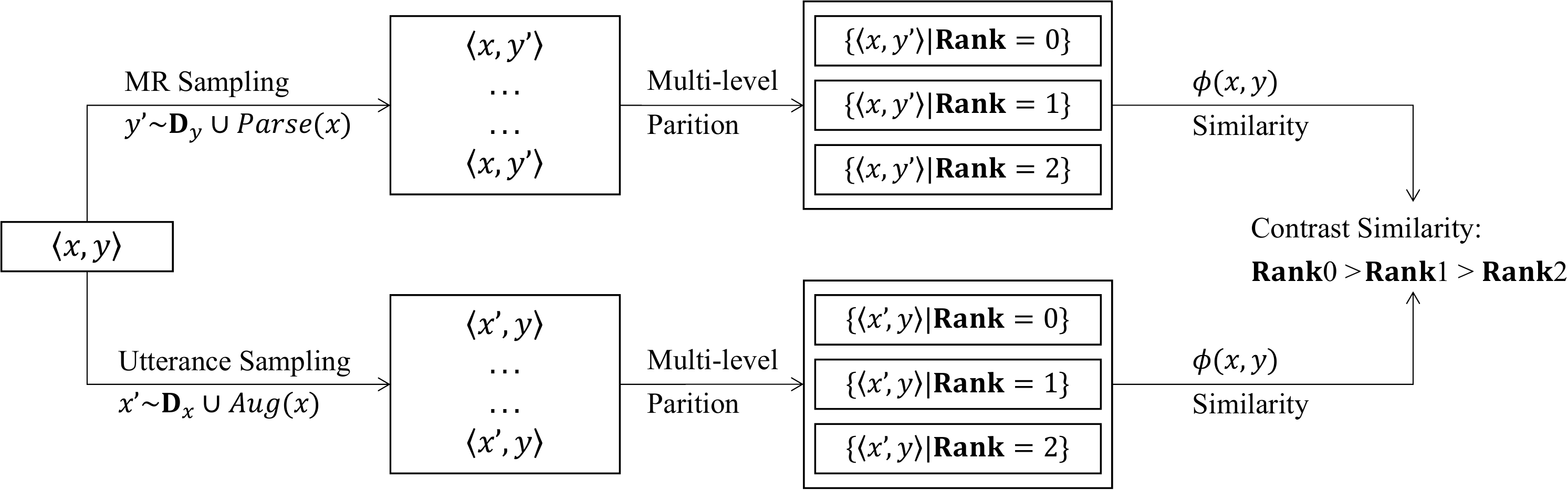} 
  \label{figbig} 

    \setlength{\abovecaptionskip}{0.3cm}
    \setlength{\belowcaptionskip}{-0.1cm}
  \caption{
The architecture of our method, in which (1) we first collect samples via online sampling where $Parse(x)$ are the parses of the current model and $Aug(x)$ are the paraphrased augmentations; (2) then we divide samples into multi-levels, where $\mathbf{Rank}\!=\!0$ indicates true positive, $\mathbf{Rank}\!=\!2$ indicates true negative, and $\mathbf{Rank}\!=\!1$ is vague instances containing paraphrases and potential aliases; (3) finally we use Ranked Contrastive Loss for optimization based on the similarities between the multi-level samples.
}
\end{figure*}


\section{Semantic-aware Contrastive Learning for Semantic Parsing}

In this section, we describe how to address the sequence-level semtantics and fine-granularity challenges via semantic-aware constrastive learning. 
The contrastive learning aims to disperse apart semantic-distinct instances and pull closer semantic-identical instances on vector representation space. 
Specifically, to learn to differentiate fine-grained meaning representations, we design a multi-level online sampling algorithm, which collects confusing negative samples and diverse positive samples in multi-level way. 
To comparing meaning representations as a whole at the sequence level, we design three semantics-aware compatibility functions. 
To learn accurate and discriminative semantic parsers, we propose \textit{ranked contrastive loss} to support the multi-level samples. 
In following we describe them in detail.

\subsection{Multi-level Online Instance Sampling}
Contrastive learning algorithms learn good parameters by trying to pull positive instances closer and push negative instances away. 
Positive and negative instances play a fundamental role in constrastive learning \citep{CLQA/Karpukhin:2020,SimCSE/Gao:2021}, and many studies focus on how to construct good positive and negative instances.

In semantic parsing, it is challenging to sample good contrastive instances. 
Firstly,  because the meaning representation is formal and diverse, it is hard to tell the the changes on semantics after a small perturbation.
Secondly, to distinguish the fine-grained semantic representations, contrastive learning needs accurate negative/positive samples. 
However, many instances are vague, which cannot be accurately categorized into positive-negative partitions. 
For example, paraphrasing, one common way to build positive instances, may changes the original fine-grained semantics. 
And two very different MRs may represent the same meaning, and cannot be treated as negative samples. 
To resolve the above challenges, we propose a multi-level partition algorithm to address vague instances, and sample instances via an online algorithm.




\subsubsection{Multi-Level Sample Partition}


In contrastive learning, each instance is a pair of utterance and MR $\left<x,y\right>$. 
To address the vagueness of instances, we divide samples into different levels according its confidence, and set each instance with a $\mathbf{Rank}$ value. 
Specifically, 
$\mathbf{Rank}\!=\!0$ indicates true positive instances, 
$\mathbf{Rank}\!=\!2$ indicates true negative instances, and 
$\mathbf{Rank}\!=\!1$ indicates vague instances which may be correct. 
In following we describe how to divide instances into these levels and leave the sampling algorithm to next section.


\paragraph{$\mathbf{Rank}\!=\!0$:} 
This level contains true positive samples. We use the golden annotations in training corpus as positive instances. And  two common types of aliases are also used as positive samples, which are show in Table \ref{aliases}. Given a MR, the utterance labeled as it and its aliases are used as positive samples.

\begin{table}[!b]
  \centering
  \resizebox{0.485\textwidth}{!}{
    \begin{tabular}{|c|l|}
    \specialrule{0em}{0pt}{12pt}
    \hline
    \fontsize{10pt}{10pt}{Types}
    &
    \tabincell{c}{\fontsize{10pt}{10pt}{$\ \!$Examples}} 
     \\
    \hline
    \renewcommand\arraystretch{0.99}
    \tabincell{c}{\specialrule{0em}{1pt}{1pt}\fontsize{10pt}{10pt}{Expressions swapping}\\ 
    \fontsize{10pt}{10pt}{in conjunction}} 
    &
    \renewcommand\arraystretch{0.99}
    \tabincell{l}{\specialrule{0em}{1pt}{1pt}\fontsize{10pt}{10pt}{$filter ( type ( meeting ) ,  attendee ,  = , concat ( alice , bob ) ) )$}\\
    \fontsize{10pt}{10pt}{$ filter ( type ( meeting ) ,  attendee ,  = , concat (  bob , alice ) ) )$}} 
     \\
    \hline
    \renewcommand\arraystretch{0.99}
    \tabincell{c}{\specialrule{0em}{1pt}{1pt} \fontsize{10pt}{10pt}{Redundant type}\\ 
    \fontsize{10pt}{10pt}{assurance}} 
    &
    \renewcommand\arraystretch{0.99}
    \tabincell{l}{ \fontsize{10pt}{10pt}{$filter (  s , assists , < , ensureNumericEntity ( 3\ assist ) ) $}\\
    \fontsize{10pt}{10pt}{$filter (  s , assists , < , 3\ assist ) $}} 
     \\
    \hline
    \end{tabular} }%
  \caption{Two common types of  MR aliases. 
  As \citet{Guo:2020} mentioned, because of the variety and composability of the aliases, there are a lot of diverse aliases of MR.
  }
  \label{aliases}%
\end{table}

\paragraph{$\mathbf{Rank}\!=\!1$:} 
This level contains vague samples which we cannot clearly identify whether it is into positive or negative. 
There are two types of vague instances. 
One is the utterance paraphrased version of instances, i.e., we paraphrase the annotated pair $\left<x,y\right>$ and obtain $\left<x',y\right>$.  
Because paraphrasing may change the original semantic, we set this instance as vague one. 
The other is the MR aliases version, i.e., for a positive instance $\left<x,y\right>$ we view $\left<x,y''\right>$ and $\left<x'',y\right>$ as vague instances if the annotated $\left<x'',y''\right>$ instance has the same execution result as $y$. 
Because the same execution result means they are potential aliases and may entail the same semantic, we use them as vague instances.

Because these samples are vague, directly adding them as positive will  mislead the model, but ignoring them may reduce diversity of positive instances and thus affect the model generalization ability. 
Therefore we view them as vague instances.



\paragraph{$\mathbf{Rank}\!=\!2$:} 
This level contains true negative instances. 
For utterance, negative MRs are the MRs with the wrong execution results.
For MR, negative utterances are the ones labeled with the MRs producing wrong execution results .

\subsubsection{Instance Sampling}

There are two common sampling algorithms for contrastive learning:

1) Batch sampling: The positive and negative sample pairs are collected from the same batch. As shown in SimCLR\citep{SimCLR/Chen:2020}, this algorithm is efficient and simple.

2) Online sampling: Given an annotated $\left<x,y\right>$ pair, we sample its positive, vague, and negative instances during parsing. Given an input utterance, we use the top-$K$  parses as candidates, and then devide them into $\mathbf{Rank}\ 0, 1, 2$ according to the methods described in above.

Because online sampling can collect hard negative samples(i.e., the top $K$ ranked instances), this paper use it for better distinguishing confusing, fine-grained meaning representations.





\subsection{Ranked Contrastive Loss}


Traditional contrastive learning only considers positive and negative samples, and their instances are usually not fine-grained. In our semantic-aware contrastive learning, we need to deal with multi-level instances, and use special semantic-aware similarities.

To this end, we propose \textit{ranked contrastive loss}, which aims to learn accurate and robust representations from the multi-level sample instances. Concretely, given sample instances in several  levels, ranked contrastive loss compare both utterances and meaning representations:
\begin{align}
\mathcal{L}_{ctrs\_{Utt}} \! & = \! \sum_{r=0,1} \sum\limits_{\scriptscriptstyle{\mathbf{Rank}_{x}= r}} \!\!\!\! -log  \frac{ e^{\phi_\theta(x,y)/\tau}}
{\sum\limits_{\scriptscriptstyle{\mathbf{Rank}_{x'} \geq r}} \!\!\!\! e^{\phi_\theta(x',y)/\tau}} & \!\!\! \!\!\! \!\!\! \\
\mathcal{L}_{ctrs\_{MR}} \! & = \! \sum_{r=0,1} \sum\limits_{\scriptscriptstyle{\mathbf{Rank}_{y}= r}} \!\!\!\!  -log \frac{ e^{\phi_\theta(x,y)/\tau}}
{\sum\limits_{\scriptscriptstyle{\mathbf{Rank}_{y'} \geq r} }\!\!\!\! e^{\phi_\theta(x,y')/\tau} }
\end{align},
in which $\tau$  denotes a temperature parameter.

When there are only positive and negative samples as two ranks, the ranked contrastive loss  can be gracefully  degraded into the ordinary InfoNCE loss \citep{CPC/Oord:2018,CPC/Carse:2021}: 
With the minibatch $\mathcal{B}$ of size $k$, consisting of one positive pair $(x, y)$ and $k-1$ negative pairs $(x^{\prime}, y)$, the InfoNCE loss is defined as
\begin{equation}
\mathcal{L}_k=\mathbb{E}_{\mathcal{B}}\left[-\log \frac{e^{\phi_{\theta}(x, y)/\tau}}{e^{\phi_{\theta}(x, y)/\tau}+\sum_{i=1}^{k-1} e^{\phi_{\theta}\left(x_{i}^{\prime}, y\right)/\tau}}\right]
\end{equation}
, which is also proved to be a lower bound on the mutual information of $x$ and $y$.

The final training objective is to minimize the decoding loss and contrastive loss as follows:
\begin{equation}
\mathcal{L} = \mathcal{L}_{seq2seq} + \alpha \mathcal{L}_{ctrs\_Utt} + \beta  \mathcal{L}_{ctrs\_MR} 
\end{equation}
where $\alpha$ and $\beta$ are hyper-parameters that represent the weights of the contrastive learning.
In this way, the model can reduce the influence of noise in sample instances and robustly improve the generalization by the augmentation of the instances.

\subsection{Semantic-aware Compatibility Function}


In contrastive learning, it is critical to measure the similarities between utterances and meaning representations, 
so that the positive $\left<x,y\right>$ instances will have high similarity, 
and the negative instances will have low similarity.
As described above, a semantic parsing system needs to measure the similarity by taking semantic representations as a whole.
Specifically, we design three compatibility functions on sequence representations, attention-based representations and MR-conditioned representations.

\paragraph{Compatibility Function on Sequence Representations}
This similarity measure takes both utterance and meaning representations as two token sequences. 
We project the embedding representations of utterances and MRs onto the latent embedding space, and obtain the similarity between them:
\begin{equation}
  \begin{aligned}
\phi_{sr}(x,y) = mean(\mathbf{h}_\mathbf{x})^TW_{s}\ mean(\mathbf{g}_\mathbf{y})
  \end{aligned}
\end{equation}
, in which $\mathbf{h}_\mathbf{x}$ is the encoded contextual embedding of  utterance $\mathbf{x}$ in \textsc{Seq2Seq} encoder, and $\mathbf{g}_\mathbf{y} = \mathbf{g}_1,\mathbf{g}_2,...,\mathbf{g}_{|y|}$ is the encoded representations of $y$. An additional LSTM encoder is employed to represent MRs, which also shares the same token embeddings with the decoder.


\paragraph{Compatibility Function with Attention}
Because different tokens may have different importances, we extend the above mean pooling with  attention mechanism as a soft selection to compute token-specific sentence representations.
\begin{gather}
 a(\mathbf{h}_{i}, \mathbf{g}_{t})  = \frac{e^{\mathbf{h}_{i}^T W_a \mathbf{g}_{t}}}{\sum_{i'=1}^{|x|}e^{\mathbf{h}_{i'}^T W_a \mathbf{g}_{t}}} \\
\tilde{\mathbf{h}}_{y_t} =\sum_{i=1}^{|x|} a\left(\mathbf{h}_{i}, \mathbf{g}_{t}\right)\;\mathbf{h}_{i}
\end{gather}
Then the compatibility function is:
\begin{gather}
\phi_{att}(x,y)= \sum_{t=1}^{|y|} \tilde{\mathbf{h}}_{y_t}^T W_{att} \mathbf{g}_t
\end{gather}

\begin{table*}[!t]
  \centering
  \resizebox{0.89\textwidth}{!}{
    \begin{tabular}{l|l|cccccccc|l}
    \Xhline{0.88pt}
    \multicolumn{2}{c|}{   }&
    \bf{Bas.} & \bf{Blo.} & \bf{Cal.} & \bf{Hou.} & \bf{Pub.} & \bf{Rec.} & \bf{Res.} & \bf{Soc.} & \bf{Avg.} \\
    \Xhline{0.88pt}
    \multicolumn{10}{l}{ \textbf{Previous} } \\
    \hline
    \multicolumn{2}{l|}{\textsc{SPO} {\cite{Wang:2015}}} 
    & 46.3 & 41.9 & 74.4 & 54.0 & 59.0 & 70.8 & 75.9 & 48.2 & 58.8\\
    \multicolumn{2}{l|}{\textsc{DSP-C } {\cite{DSP/Xiao:2016}}} 
    &80.5 &55.6 &75.0 &61.9 &75.8 & - &80.1 &80.0 &72.7\\
    \multicolumn{2}{l|}{\textsc{Recomb} {\cite{Robin:2016}}} & 
85.2 & 58.1 & 78.0 & 71.4 & 76.4 & 79.6 & 76.2 & 81.4 & 75.8 \\
    \multicolumn{2}{l|}{\textsc{Recomb} {\cite{Robin:2016}}(+data)}
   & 87.5 & 60.2 & 81.0 & 72.5 & 78.3 & 81.0 & 79.5 & 79.6 & 77.5 \\
    \multicolumn{2}{l|}{\textsc{CrossDomain} {\citep{Su:2017}}} & 86.2 & 60.2 & 79.8 & 71.4 & 78.9 & 84.7 & 81.6 & 82.9 & 78.2 \\
    \multicolumn{2}{l|}{\textsc{\textsc{Seq2Action}} {\citep{Chen:2018}}} & 
    \textbf{88.2} & 61.4 & 81.5 & 74.1 & 80.7 & 82.9 & 80.7 & 82.1 & 79.0 \\
    \multicolumn{2}{l|}{\textsc{Dual} {\citep{Cao:2019}}} & 87.0 & \textbf{66.2} & 79.8 & 75.1& 80.7& 83.3& 83.4& 83.8& 79.9 \\
    \multicolumn{2}{l|}{\textsc{Two-stage} {\citep{Cao:2020}}} & 87.2 & 65.7 & 80.4 & 75.7 & 80.1 & \textbf{86.1} & 82.8 & 82.7 & 80.1 \\
    \multicolumn{2}{l|}{\textsc{SSD} {\citep{SSD/Wu:2021}}} & 86.2 & 64.9 & 81.7 & 72.7 & 82.3 & 81.7 & 81.5 & 82.7 & 79.2  \\
    \Xhline{0.88pt}
    \multicolumn{10}{l}{ \textbf{Our Methods} } \\
    \hline
    \multicolumn{2}{l|}{\textsc{Seq2Seq}}
    &86.4  & 61.4 & 73.8 & 68.3 & 76.4 & 77.8 & 78.3 & 82.8 & 75.7  \\
    \hline
    \multicolumn{2}{l|}{{SemCL}  (SR)}
    & 88.0 &  64.7 & 81.0& 77.8& 80.7 & 84.7 & 84.0 & 83.0 & 80.4 \\
    \multicolumn{2}{l|}{{SemCL}  (Att)}
    & 87.7 &  65.2  & 81.5& 76.2& 80.7 & 82.4  & 83.7 & 83.8 & 80.2  \\
    \multicolumn{2}{l|}{{SemCL}  (Cond)}
     & \textbf{88.2} &  65.7 & \textbf{82.7}& \textbf{78.3}& \textbf{81.4} & 83.8  & \textbf{84.3}  & \textbf{84.2} & \textbf{81.1 } \\
    \hline

    \Xhline{1pt}
    \end{tabular} }%
    \setlength{\abovecaptionskip}{0.2cm}
    \setlength{\belowcaptionskip}{-0.3cm}
  \caption{Overall results on \textsc{Overnight}.
  }
  \label{Overnight}%
\end{table*}%

\paragraph{MR-Conditioned Compatibility Function}

Semantic parsing is a \textsc{Seq2Seq} generation process, and the decoder decides which utterance tokens are used to decode a MR token $y_t$. 
Therefore we take these conditional association into consideration, and measure the similarity between $x$ and $y$. 
In \textsc{Seq2Seq} decoding, $c_t$ is the attentioned source context representation in the decoding step as in Equ.  \ref{ct}.
Then the compatibility function is:
\begin{gather}
\phi_{cond}(x,y)= \sum_{t=1}^{|y|} c_t^T W_{c} \mathbf{g}_t
\end{gather},
where $c_t$ captures the used parts of utterance representation in decoding.




\section{Experiments}

\subsection{Experimental Settings}
\paragraph{Datasets}

We conduct experiments on \textsc{Overnight} and \textsc{GeoGranno}, which involve various domains. Our implementations are public available$\footnote{\text{https://github.com/lingowu/semcl}}$.

\indent \textbf{\textsc{Overnight}}\indent  
This is a multi-domain dataset, which contains natural language queries paired with lambda DCS logical forms. 
The \textsc{Overnight} benchmark consists of eight semantic parsing datasets covering a range of semantic phenomena, 
which requires precise semantics learning ability to map natural language queries to the structured logical forms.
We  use the same train/test splits as \citet{Wang:2015} to choose the best model during training.


\indent \textbf{\textsc{GeoGranno}}\indent  
This is an version of \textsc{Geo} \citep{Zelle:1996,Herzig:2019}, which is labeled with lambda DCS logical forms.
The dataset is constructed by paraphrases detecting.
Crowd workers are employed to select the correct canonical utterance from candidate list.
The generalization ability of models are requisite to handle  278 test queries from small numbers of train examples with only 487  instances.
We follow the same splits in original paper \citep{Herzig:2019}. 

In all our experiments, the standard accuracy is used to evaluate systems. 
The accuracies on all datasets are  calculated as the same as \citet{Robin:2016} and \citet{Herzig:2019}.

\paragraph{Data Preprocessing} 
Following \citet{Dong:2016}, we handle entities with Replacing mechanism, which replaces identified entities with their types and IDs. 
The entity mapping lexicons in \citet{Cao:2019} are also used.
The paraphrasing model is the trained paraphraser based on T5$\footnote{http://github.com/PrithivirajDamodaran/Parrot\_\newline Paraphraser}$, and  we paraphrase 20 different expressions for each utterances.

\paragraph{System Settings} 
The bidirectional LSTM encoders are employed for utterances and MRs with  200 hidden units and 300-dimensional word vectors. 
We also use 200 hidden units and 300-dimensional word vectors for LSTM decoders.
We initialize all parameters by uniformly sampling within the interval [-0.1, 0.1].
The batch size is set as 128. 
For each MR/utterance we collect 5 paraphrases and 100 random utterance/MR samples.
In online sampling, after each training epoch, we collect the additional samples from the beam search results with the beam size as 20.
We take $\alpha\!=\!1$, $\beta\!=\!1$, and $\tau\!=\!0.3$.
The first 5 epochs train the original model, and the following 25 epochs optimize the overall training loss. 
The beam size of the decoder is set to 10. 
We use optimizer Adam\citep{Adam/Kingma:2015}  with learning rate 0.001 for all experiments.
In the main experiments and ablation experiments, our models use online sampling and multi-level partition by default.

\subsection{Experimental Results}

\subsubsection{Overall Results}

The overall results of baselines and different settings of our method are shown in Table \ref{Overnight} and Table \ref{Geo}. 
SR, Att, and Cond indicate the above three compatibility functions.
We can see that:

1. \textbf{Semantic-aware contrastive learning is effective, which achieves state-of-the-art performance using a simple base model.} 
On \textsc{Overnight} and \textsc{GEOGranno} dataset, we both achieve state-of-the-art performance on average (81.1\% and 73.4\%). 
The results demonstrate the superiority of our contrastive learning algorithms.

2. \textbf{By taking the fine granularity and sequence-level semantics  into consideration, the semantic-aware contrastive learning can significantly outperform MLE algorithm.} 
Compared with the MLE counterpart -- \textsc{Seq2Seq}, 
the contrastive learning algorithm can lead to 5.4 and 1.8 accuracy improvements on \textsc{Overnight} and \textsc{GeoGranno}. 
This verifies that compared with MLE, our semantic-aware constrastive learning can learn more accurate semantic parsers.

3. \textbf{Semantic-aware similarity is critical for accurate semantic parsing.}
We can see that, all compatibility functions show their advantages over MLE-baselines. 
And more accurate similarity measure can result better performance. 
Such as, MR-conditioned compatibility functions are more stable in various domains and datasets. 
In general, the improvement of using semantic-aware contrastive learning is significant, regardless of which function in the three ones is used.   

\begin{table}[!t]
  \centering
  \resizebox{0.395\textwidth}{!}{
    \begin{tabular}{l|l|c}
    \Xhline{0.8pt}
    \multicolumn{3}{l}{ \textbf{Previous Methods} } \\
    \hline
    \multicolumn{2}{l|}{\textsc{CopyNet} \small{\citep{Herzig:2019}}} & 72.0 \\
    \multicolumn{2}{l|}{One-stage \small{\citep{Cao:2020}}} & 71.9 \\
    \multicolumn{2}{l|}{Two-stage \small{\citep{Cao:2020}}} & 71.6 \\
    \multicolumn{2}{l|}{\textsc{SSD }Word-Level \small{\citep{SSD/Wu:2021}}} &  72.9  \\
    \multicolumn{2}{l|}{\textsc{SSD }Grammar-Level \small{\citep{SSD/Wu:2021}}} & 72.0  \\
    \Xhline{0.8pt}
    \multicolumn{3}{l}{ \textbf{Our Methods} } \\
    \hline
    \multicolumn{2}{l|}{\textsc{Seq2Seq}} &  71.6   \\
    \hline
    \multicolumn{2}{l|}{SemCL (SR)} &  73.0  \\
    \multicolumn{2}{l|}{SemCL (Att)} &   72.6 \\
    \multicolumn{2}{l|}{SemCL (Cond)} & 73.4   \\
    \Xhline{0.85pt}
    \end{tabular} }%
    \setlength{\abovecaptionskip}{0.2cm}
    \setlength{\belowcaptionskip}{-0.3cm}
  \caption{Overall results on \textsc{GeoGranno}.
  }
  \label{Geo}%
\end{table}%

\subsubsection{Detailed Analysis}

\begin{table*}[h]
  \centering
  \resizebox{0.85\textwidth}{!}{
    \begin{tabular}{lccccccccccc}
    \toprule
    & \bf{Bas.} & \bf{Blo.} & \bf{Cal.} & \bf{Hou.} & \bf{Pub.} & \bf{Rec.} & \bf{Res.} & \bf{Soc.} 
    & 
    \bf{Avg.} 
    &
    \\
    \hline    
    \specialrule{0em}{0pt}{1pt}
    \textsc{FullModel}   
    & \textbf{88.2} &  \textbf{65.7} & \textbf{82.7}& \textbf{78.3}& \textbf{81.4} & \textbf{83.8}  & \textbf{84.3}  & \textbf{84.2}& \textbf{81.1} 
      &   \\ 
    \hline
    \specialrule{0em}{0pt}{1pt}
    \multicolumn{12}{l}{ \textbf{Multi-Level Samples} } \\
    \hline
    $\ \ $   $Rank=1$ neglect &  87.7 &  64.7 & 79.8 & 75.7 & 79.5  & 82.4 & 83.7 & 83.9  & 79.7  &   &   \\
    $\ \ $ $Rank=1$ as positive  & \textbf{88.2} & 63.7 & 81.0 & 75.1 & 78.3 & 81.0 & 80.4 & 82.9 & 78.8  &  &   \\ 
    $\ \ $ $Rank=1$ as negative  & 86.2 & 63.2 & 75.0 & 72.0 & 77.6 & 79.6 & 81.3 & 83.0 & 77.2    
     \\ 
    \hline
    \specialrule{0em}{0pt}{1pt}
    \multicolumn{12}{l}{ \textbf{Precise Sampling} } \\
    \hline

    $\ \ $ Batch Sampling
     & 87.5 & 63.7 & 80.4 & 75.1 & 78.3 & 82.9 & 82.5 & 82.6 & 79.1 
   \\ 
    $\ \ $ Random Sampling
 & 87.2 & 64.4 & 81.0 & 74.6 & 79.5 & 82.4 & 81.6 & 83.6 & 79.3 
     \\ 
    \hline
    \specialrule{0em}{0pt}{1pt}
    \multicolumn{12}{l}{ \textbf{Contrastive Loss} } \\
    \hline
    $\ \ $ {\textsc{SemCL}  ($\mathcal{L}_{\scriptstyle{ctrs\_MR}}$ \& SR)}
    & 87.5& 62.2& 81.0& 71.4& 77.0& 82.4 & 80.4& 82.8& 78.1  \\
    $\ \ $ {\textsc{SemCL}  ($\mathcal{L}_{\scriptstyle{ctrs\_MR}}$ \& Att)}
    & 87.2& 63.7& 80.4& 73.5& 78.3& 81.5 & 78.9& 83.3& 78.3  \\
    $\ \ $ {\textsc{SemCL}  ($\mathcal{L}_{\scriptstyle{ctrs\_MR}}$ \& Cond)}
    & 87.7& 62.7& 81.5& 74.1& 78.9& 81.0 & 79.2& 83.6& 78.6 \\
    \hline
    $\ \ $ {\textsc{SemCL}  ($\mathcal{L}_{\scriptstyle{ctrs\_Utt}}$ \& SR)}
    & 86.7 &  64.4  &  81.0 & 75.7  & 77.6  & 81.5  & 83.4  & 82.5  &  79.1  \\
    $\ \ $ {\textsc{SemCL}  ($\mathcal{L}_{\scriptstyle{ctrs\_Utt}}$ \& Att)}
    & 86.2 &  64.9 &  79.2 &  74.6 & 78.3  &  81.9 & 82.8 & 83.3  &  78.9  \\
    $\ \ $ {\textsc{SemCL}  ($\mathcal{L}_{\scriptstyle{ctrs\_Utt}}$ \& Cond)}
     & 87.0 &  65.2 &  81.0 &  76.2 &  78.9 & 81.5  & 83.1 & 83.4  &  79.5  \\
    \bottomrule
    \end{tabular}  }%
    \setlength{\abovecaptionskip}{0.2cm}
    \setlength{\belowcaptionskip}{-0.2cm}
  \caption{ Ablation results of our model with different settings on \textsc{Overnight}.
  }
  \label{abaltion_all}%
\end{table*}%

\begin{figure*}[!t]
  \centering
    \subfigure[\textsc{Seq2Seq}]{
      \begin{minipage}[t]{0.297\textwidth}
        \centering
        \includegraphics[width=0.98\textwidth]{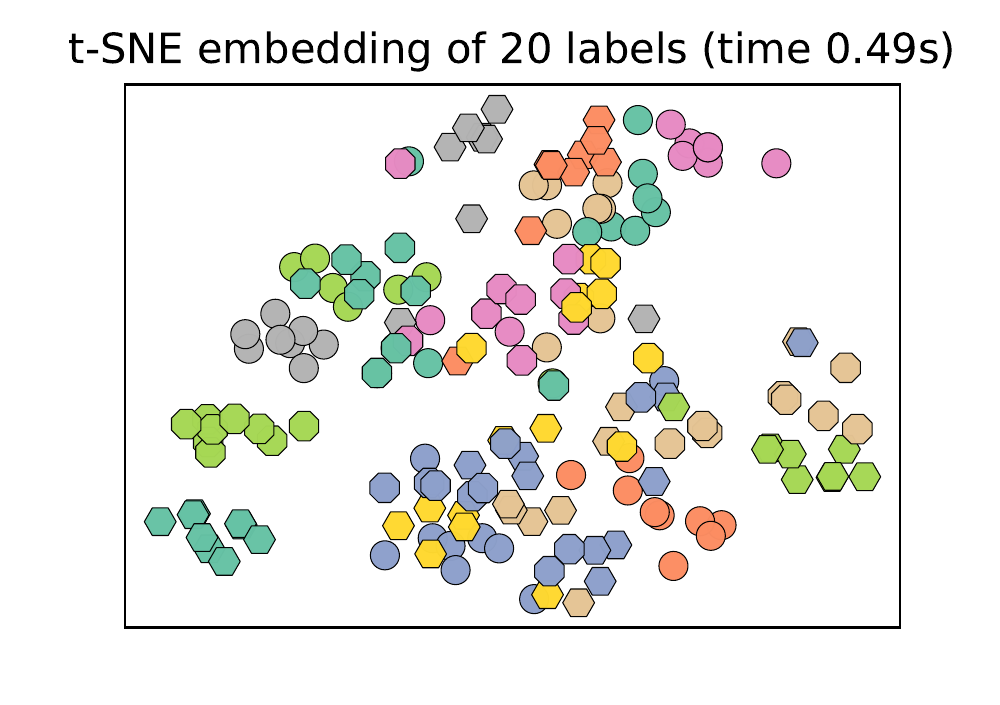} 
      \label{tSNE-1} 
      \end{minipage}
    } $\ \!\!\!\!\!\;$
    \subfigure[SemCL(without paraphrases)]{
      \begin{minipage}[t]{0.297\textwidth}
      \centering
      \includegraphics[width=0.9785\textwidth]{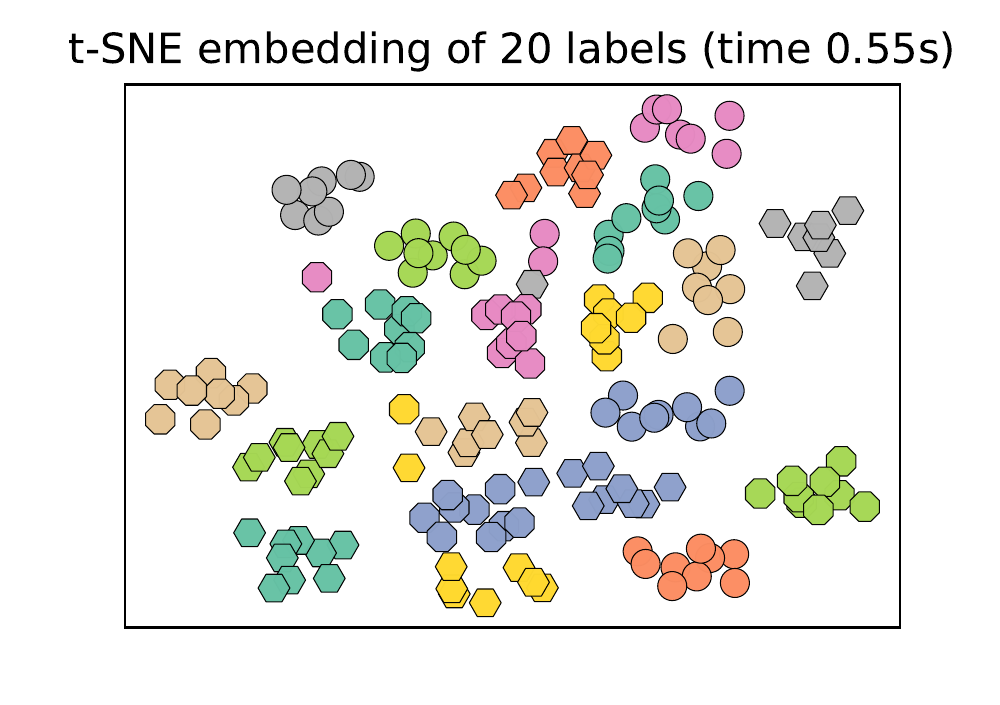} 
      \label{tSNE-2} 
      \end{minipage}%
    } $\ \!\!$
    \subfigure[SemCL(paraphrases as $\mathbf{Rank}\!\!=\!\!1$) ]{
      \begin{minipage}[t]{0.297\textwidth}
      \centering
      \includegraphics[width=0.98\textwidth]{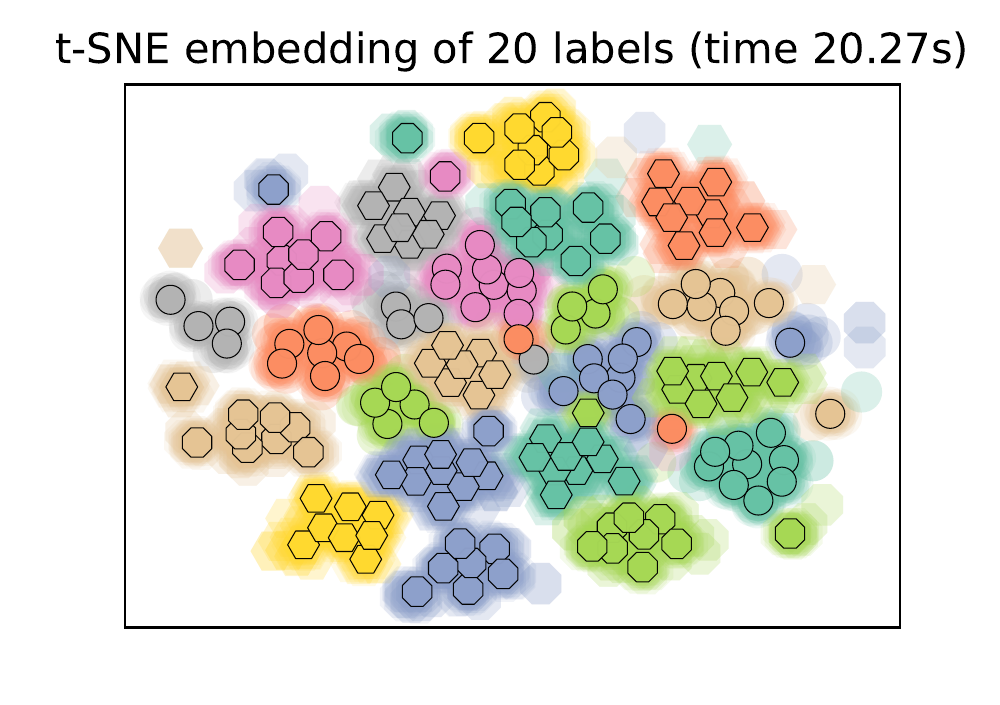} 
      \label{tSNE-3} 
      \end{minipage}%
    }
    \setlength{\abovecaptionskip}{0.1cm}
    \setlength{\belowcaptionskip}{-0.3cm}
  \caption{
Comparison of t-SNE visualization for the learned utterance representations. The colors and markers indicate different MR labels. The transparent markers indicate the representations of the paraphrases.}
  \label{tSNE} 
\end{figure*}

\paragraph{Effect of multi-level sampling}
To analyze the effect of the multi-level samples partition, we conduct experiments with the positive-negative samples partition. The results are shown in Table \ref{abaltion_all}.
When there are only positive and negative samples, we try three ways to deal with the part of $\mathbf{Rank}\!=\!1$: ignoring it or viewing it as positive or negative samples.
We can see that treating it as negative is inadvisable, which brings significant performance degradation. We think this is because there are many positive examples in $\mathbf{Rank}\!=\!1$ part, which should be gathered in representation space. And viewing them as negatives will mislead the model learning.
We can see that treating it as positive also brings slight performance drop, which may be due to the noise samples.
The results also show that our approach is better than neglecting it. We believe that ignoring them will lead to insufficient generalization ability of the model.
In general, it is problematic to employ vague instances in positive-negative partition, and our multi-level fashion can facilitate the learning of vague samples.

\paragraph{Effect of contrastive losses} 
To investigate the effect of contrastive losses, we compare the settings with only contrastive learning on utterance side or MR side.
The results in Table \ref{abaltion_all} show the performances of using contrastive losses on both sides are the best in all domains.
We believe that by jointly optimizing the constrastive losses in both sides, the model can learn better sentence-side and semantic-side representations, which are beneficial to the model's awareness of fine-grained semantics.

\paragraph{Effect of  instances sampling} 
We conduct another experiment by changing sampling methods  and the results are shown in Table \ref{abaltion_all}: Batch Sampling denotes the sample instances are collected in the same batch; Random Sampling denotes for each utterance/MR we randomly select 100 MRs/utterances to form contrastive samples.
We can see the instance-level sampling is important. The Random Sampling on the instance-level is slightly better than Batch Sampling. 
We can see that the performances of Online Sampling (\textsc{FullModel}) is significantly higher than other methods.
This verifies that differentiating the hard negative samples, which are confusing on the current model, makes the model learn the fine-grained semantics more accurately.

\paragraph{Visualization of the representations space of utterances}
We use t-SNE \citep{tSNE2008} to visualize the utterance representations on a 2D map. 
The utterance representations are calculated by averaging the hidden states on words.
The encoders of \textsc{Seq2Seq} models and our contrastive learning models(with sequence representations compatibility function)  are used to obtain utterance representations.
In Fig \ref{tSNE}, we draw the most frequent 20 MRs and their utterances.
Compared with \textsc{Seq2Seq} baseline, our two models learn smoother representation space, which explains why SemCL can yield better parsing performance.
In Fig \ref{tSNE-3}, we plot the 20 paraphrases for each utterance with the transparent markers.
Although there are some noise samples in the paraphrases, compared with no paraphrases, the generalization ability of the model is improved by our multi-level sampling mechanism.

\begin{table}[!t]
  \centering
  \resizebox{0.49\textwidth}{!}{
    \begin{tabular}{|l|l|}
    \hline
    \multicolumn{2}{|l|}{
    \tabincell{l}{ 
    \specialrule{0em}{0pt}{3pt}
    \textbf{Input}: articles published 2004 or more recent}}  \\
    \multicolumn{2}{|l|}{
    \tabincell{l}{ 
    \specialrule{0em}{0pt}{3pt}
    \textbf{Baselines}: $listValue\ ( filter\ ( type\ ( article ) , publication\_date,$ \\ $ {\color{red}= , cat\ (date\_2004,date\_2010) }) ) $
    } }  \\
    \multicolumn{2}{|l|}{
    \tabincell{l}{ 
    \specialrule{0em}{0pt}{3pt}
    \textbf{SemCL}: $listValue\ ( filter\ ( type\ ( article ) , publication\_date,$ \\ 
    $ \color{blue}{\ge , date\_2004} ) )$
    }} \\
    \multicolumn{2}{|c|}{
    \tabincell{l}{ 
    (a)  domain: \textsc{Publication} \\
    \specialrule{0em}{0pt}{2pt}}} \\
    \hline
    \multicolumn{2}{|l|}{
    \tabincell{l}{ 
    \specialrule{0em}{0pt}{3pt}
    \textbf{Input}: how many points did kobe bryant have when he had 3 block}} \\
    \multicolumn{2}{|l|}{
    \tabincell{l}{ 
    \specialrule{0em}{0pt}{3pt}
    \textbf{Baselines}: $listValue\ ( property\ ( filter\ ( property\ ( kobe\_bryant , $\\ 
    $reverse ( player ) ) , {\color{red}num\_points} , = , 3\ block ) , num\_points ) ) $
    }} \\
    \multicolumn{2}{|l|}{
    \tabincell{l}{ 
    \specialrule{0em}{0pt}{3pt}
    \textbf{SemCL}: $listValue\ ( property\ ( filter\ ( property\ ( kobe\_bryant , $\\ 
    $reverse ( player ) ) , {\color{blue}num\_blocks} , = , 3\ block ) , num\_points ) ) $
    }} \\
    \multicolumn{2}{|c|}{
    \tabincell{l}{ 
    (b)  domain: \textsc{Basketball} \\
    \specialrule{0em}{0pt}{2pt}}} \\
    \hline
    \end{tabular} }%
    \setlength{\abovecaptionskip}{0.1cm}
    \setlength{\belowcaptionskip}{-0.4cm}
  \caption{Cases on \textsc{Overnight} with MLE baseline and our  SemCL.
  The MRs are simplified  for readability.
  }
  \label{cases}%
\end{table}%

\paragraph{Case study}
In Table \ref{cases} we compare the parsed results of our model with that of the \textsc{Seq2Seq} baseline. 
In domain \textsc{Publication}, 
the utterance is mistaken by baseline for similar semantic ``articles published in 2004 or 2010''.
Our model can well distinguish the fine-grained difference and generate the correct MR, even the expression about ``recent'' is rare in the training set. We think both generalization and discrimination of the models are improved.
In domain \textsc{Basketball}, the baseline  over-translate the word ``points''. But SemCL can perceive the whole semantics and produce the right MR, which shows the effectiveness of treating both the utterance and the MR as a whole.

\section{Related Work}

\paragraph{Contrastive Learning.}
Contrastive learning is a method of representation learning \citep{CL/Hadsell:2006}, which pulls the relevant embeddings together and pushes different ones apart to  provide more effective representations \citep{CPC/Oord:2018,SimCLR/Chen:2020}.
In NLP, contrastive learning is widely used in sentence representations learning \citep{CODA/Qu:2021,SimCSE/Gao:2021,CL-BERT/Kim:2021,CL-SR/Giorgi:2021}. Contrastive learning also improves many natural language understanding tasks \citep{CIL/Chen:2021,CLINE/Wang:2021,CLEVE/Wang:2021,ERICA/Qin:2021}.
To the best of our knowledge, our work is  the first attempt to adopt contrastive learning for semantic parsing.

\paragraph{Sequence-level Semantics.}
In \textsc{Seq2Seq} tasks, many studies have been proposed to remedy the problem of MLE, such as, minimum risk training \citep{MinRisk/Shen:2016}, contrastive approaches \citep{CL4Gen/Lee:2021} and reinforcement learning \citep{Xia:2016,RL-para/Li:2018}.
In semantic parsing, maximum marginal likelihood methodes are proposed to exploit  consistent logical forms \citep{onlineMML/Berant:2013,onlineMML/Guu:2017,onlineMML/Goldman:2018}.
The structured learning methods are employed to maximize the margins or minimize the expected risks \citep{structuredLearning:/Yih:2015,StructVAE/Yin:2018,structuredLearning:/Xiao:2016,structuredLearning:/Iyyer:2017}.
Dual learning methods have been proposed, which also design and optimze rewards in sequence level \citep{Cao:2019,Cao:2020}.
Globally normalized models are proposed to relieve the label bias problem in MLE \citep{Global/Huang:2021}.
Different from previous works, our approach aims to acquire more discriminative sequence representations by contrastive learning.

\paragraph{Semantic Generalization.}
Recently, generalization problem has become a research hotspot in semantic parsing. 
To generalize on various natural language expressions, semantic-invariance knowledge are introduced by paraphrasing \citep{Berant:2014,Wang:2015,Dong:2017,Herzig:2019}.
There are also many strudies for achieving generalization on meaning composition \cite{CompositionalGr/Oren:2020,CompositionalGr/Liu:2020,CompositionalGr/Conklin:2021,CompositionalGr/Bogin:2021,CGIRSP/Herzig:2021}.
To generalize in low resources settings, data augmentation \citep{Robin:2016,unNLP/Marzoev:2020} and constrained decoding\citep{SSD/Wu:2021,FewShotSP/Shin:2021} methods are proposed.
In this paper, the generalization of representations is improved by pulling the utterance and MRs with similar overall semantics closer in representation space.

\section{Conclusions}

This paper proposes Semantic-aware Contrastive Learning -- an effective contrastive learning framework for semantic parsing,
which  takes the sequence-level semantics and the fine-granularity into consideration.
Specifically, we propose a multi-level online sampling algorithm for accurately collecting confusing and diverse samples, 
and design three semantic-aware similarity functions to measure the similarity between utterances and MRs.
We also propose Ranked Contrastive Loss to optimze the representations from the multi-level samples.
Experimental results show that our approach can achieve state-of-the-art performance in several domains and datasets.

\section{Limitations}
There are two main limitations of this work.
1) Since additional negative instances are used for contrasting in our SemCL method, our method requires more training time than vanilla semantic parsing methods.
2) Our proposed method still relies on an amount of annotated data. Many contrastive learning  methods have been proposed to resolve the low-resource tasks.  
We will exploit contrastive learning for low-resource semantic parsing in the future.

\section*{Acknowledgments}
We sincerely thank the reviewers for their insightful comments and valuable suggestions. 
Moreover, this work is supported by the National Natural Science Foundation of China under Grants no. U1936207, 62122077, 61906182 and 62076233.

\bibliography{SP}

\begin{thebibliography}{57}
\expandafter\ifx\csname natexlab\endcsname\relax\def\natexlab#1{#1}\fi

\bibitem[{Berant et~al.(2013)Berant, Chou, Frostig, and
  Liang}]{onlineMML/Berant:2013}
Jonathan Berant, Andrew Chou, Roy Frostig, and Percy Liang. 2013.
\newblock \href {https://aclanthology.org/D13-1160/} {Semantic parsing on
  freebase from question-answer pairs}.
\newblock In \emph{Proceedings of the 2013 Conference on Empirical Methods in
  Natural Language Processing, {EMNLP} 2013, 18-21 October 2013, Grand Hyatt
  Seattle, Seattle, Washington, USA, {A} meeting of SIGDAT, a Special Interest
  Group of the {ACL}}, pages 1533--1544. {ACL}.

\bibitem[{Berant and Liang(2014)}]{Berant:2014}
Jonathan Berant and Percy Liang. 2014.
\newblock \href {https://doi.org/10.3115/v1/p14-1133} {Semantic parsing via
  paraphrasing}.
\newblock In \emph{Proceedings of the 52nd Annual Meeting of the Association
  for Computational Linguistics, {ACL} 2014, June 22-27, 2014, Baltimore, MD,
  USA, Volume 1: Long Papers}, pages 1415--1425.

\bibitem[{Bogin et~al.(2021)Bogin, Subramanian, Gardner, and
  Berant}]{CompositionalGr/Bogin:2021}
Ben Bogin, Sanjay Subramanian, Matt Gardner, and Jonathan Berant. 2021.
\newblock \href {https://transacl.org/ojs/index.php/tacl/article/view/2489}
  {Latent compositional representations improve systematic generalization in
  grounded question answering}.
\newblock \emph{Trans. Assoc. Comput. Linguistics}, 9:195--210.

\bibitem[{Cao et~al.(2019)Cao, Zhu, Liu, Li, and Yu}]{Cao:2019}
Ruisheng Cao, Su~Zhu, Chen Liu, Jieyu Li, and Kai Yu. 2019.
\newblock \href {https://www.aclweb.org/anthology/P19-1007/} {Semantic parsing
  with dual learning}.
\newblock In \emph{Proceedings of the 57th Conference of the Association for
  Computational Linguistics, {ACL} 2019, Florence, Italy, July 28- August 2,
  2019, Volume 1: Long Papers}, pages 51--64.

\bibitem[{Cao et~al.(2020)Cao, Zhu, Yang, Liu, Ma, Zhao, Chen, and
  Yu}]{Cao:2020}
Ruisheng Cao, Su~Zhu, Chenyu Yang, Chen Liu, Rao Ma, Yanbin Zhao, Lu~Chen, and
  Kai Yu. 2020.
\newblock \href {https://www.aclweb.org/anthology/2020.acl-main.608/}
  {Unsupervised dual paraphrasing for two-stage semantic parsing}.
\newblock In \emph{Proceedings of the 58th Annual Meeting of the Association
  for Computational Linguistics, {ACL} 2020, Online, July 5-10, 2020}, pages
  6806--6817. Association for Computational Linguistics.

\bibitem[{Carse et~al.(2021)Carse, Carey, and McKenna}]{CPC/Carse:2021}
Jacob Carse, Frank~A. Carey, and Stephen~J. McKenna. 2021.
\newblock \href {https://doi.org/10.1109/ISBI48211.2021.9434140} {Unsupervised
  representation learning from pathology images with multi-directional
  contrastive predictive coding}.
\newblock In \emph{18th {IEEE} International Symposium on Biomedical Imaging,
  {ISBI} 2021, Nice, France, April 13-16, 2021}, pages 1254--1258. {IEEE}.

\bibitem[{Chen et~al.(2018)Chen, Sun, and Han}]{Chen:2018}
Bo~Chen, Le~Sun, and Xianpei Han. 2018.
\newblock \href {https://aclanthology.info/papers/P18-1071/p18-1071}
  {Sequence-to-action: End-to-end semantic graph generation for semantic
  parsing}.
\newblock In \emph{Proceedings of the 56th Annual Meeting of the Association
  for Computational Linguistics, {ACL} 2018, Melbourne, Australia, July 15-20,
  2018, Volume 1: Long Papers}, pages 766--777.

\bibitem[{Chen et~al.(2021)Chen, Shi, Tang, Chen, Wu, and
  Zhuang}]{CIL/Chen:2021}
Tao Chen, Haizhou Shi, Siliang Tang, Zhigang Chen, Fei Wu, and Yueting Zhuang.
  2021.
\newblock \href {https://doi.org/10.18653/v1/2021.acl-long.483} {{CIL:}
  contrastive instance learning framework for distantly supervised relation
  extraction}.
\newblock In \emph{Proceedings of the 59th Annual Meeting of the Association
  for Computational Linguistics and the 11th International Joint Conference on
  Natural Language Processing, {ACL/IJCNLP} 2021, (Volume 1: Long Papers),
  Virtual Event, August 1-6, 2021}, pages 6191--6200. Association for
  Computational Linguistics.

\bibitem[{Chen et~al.(2020)Chen, Kornblith, Norouzi, and
  Hinton}]{SimCLR/Chen:2020}
Ting Chen, Simon Kornblith, Mohammad Norouzi, and Geoffrey~E. Hinton. 2020.
\newblock \href {http://proceedings.mlr.press/v119/chen20j.html} {A simple
  framework for contrastive learning of visual representations}.
\newblock In \emph{Proceedings of the 37th International Conference on Machine
  Learning, {ICML} 2020, 13-18 July 2020, Virtual Event}, volume 119 of
  \emph{Proceedings of Machine Learning Research}, pages 1597--1607. {PMLR}.

\bibitem[{Conklin et~al.(2021)Conklin, Wang, Smith, and
  Titov}]{CompositionalGr/Conklin:2021}
Henry Conklin, Bailin Wang, Kenny Smith, and Ivan Titov. 2021.
\newblock \href {https://doi.org/10.18653/v1/2021.acl-long.258} {Meta-learning
  to compositionally generalize}.
\newblock In \emph{Proceedings of the 59th Annual Meeting of the Association
  for Computational Linguistics and the 11th International Joint Conference on
  Natural Language Processing, {ACL/IJCNLP} 2021, (Volume 1: Long Papers),
  Virtual Event, August 1-6, 2021}, pages 3322--3335. Association for
  Computational Linguistics.

\bibitem[{Devlin et~al.(2019)Devlin, Chang, Lee, and Toutanova}]{Devlin:2019}
Jacob Devlin, Ming{-}Wei Chang, Kenton Lee, and Kristina Toutanova. 2019.
\newblock \href {https://www.aclweb.org/anthology/N19-1423/} {{BERT:}
  pre-training of deep bidirectional transformers for language understanding}.
\newblock In \emph{Proceedings of the 2019 Conference of the North American
  Chapter of the Association for Computational Linguistics: Human Language
  Technologies, {NAACL-HLT} 2019, Minneapolis, MN, USA, June 2-7, 2019, Volume
  1 (Long and Short Papers)}, pages 4171--4186.

\bibitem[{Dong and Lapata(2016)}]{Dong:2016}
Li~Dong and Mirella Lapata. 2016.
\newblock \href {http://aclweb.org/anthology/P/P16/P16-1004.pdf} {Language to
  logical form with neural attention}.
\newblock In \emph{Proceedings of the 54th Annual Meeting of the Association
  for Computational Linguistics, {ACL} 2016, August 7-12, 2016, Berlin,
  Germany, Volume 1: Long Papers}.

\bibitem[{Dong et~al.(2017)Dong, Mallinson, Reddy, and Lapata}]{Dong:2017}
Li~Dong, Jonathan Mallinson, Siva Reddy, and Mirella Lapata. 2017.
\newblock \href {https://doi.org/10.18653/v1/d17-1091} {Learning to paraphrase
  for question answering}.
\newblock In \emph{Proceedings of the 2017 Conference on Empirical Methods in
  Natural Language Processing, {EMNLP} 2017, Copenhagen, Denmark, September
  9-11, 2017}, pages 875--886.

\bibitem[{Gao et~al.(2021)Gao, Yao, and Chen}]{SimCSE/Gao:2021}
Tianyu Gao, Xingcheng Yao, and Danqi Chen. 2021.
\newblock \href {http://arxiv.org/abs/2104.08821} {Simcse: Simple contrastive
  learning of sentence embeddings}.
\newblock \emph{CoRR}, abs/2104.08821.

\bibitem[{Giorgi et~al.(2021)Giorgi, Nitski, Wang, and
  Bader}]{CL-SR/Giorgi:2021}
John~M. Giorgi, Osvald Nitski, Bo~Wang, and Gary~D. Bader. 2021.
\newblock \href {https://doi.org/10.18653/v1/2021.acl-long.72} {Declutr: Deep
  contrastive learning for unsupervised textual representations}.
\newblock In \emph{Proceedings of the 59th Annual Meeting of the Association
  for Computational Linguistics and the 11th International Joint Conference on
  Natural Language Processing, {ACL/IJCNLP} 2021, (Volume 1: Long Papers),
  Virtual Event, August 1-6, 2021}, pages 879--895. Association for
  Computational Linguistics.

\bibitem[{Goldman et~al.(2018)Goldman, Latcinnik, Nave, Globerson, and
  Berant}]{onlineMML/Goldman:2018}
Omer Goldman, Veronica Latcinnik, Ehud Nave, Amir Globerson, and Jonathan
  Berant. 2018.
\newblock \href {https://doi.org/10.18653/v1/P18-1168} {Weakly supervised
  semantic parsing with abstract examples}.
\newblock In \emph{Proceedings of the 56th Annual Meeting of the Association
  for Computational Linguistics, {ACL} 2018, Melbourne, Australia, July 15-20,
  2018, Volume 1: Long Papers}, pages 1809--1819. Association for Computational
  Linguistics.

\bibitem[{Guo et~al.(2020)Guo, Liu, Lou, Li, Liu, Xie, and Liu}]{Guo:2020}
Jiaqi Guo, Qian Liu, Jian{-}Guang Lou, Zhenwen Li, Xueqing Liu, Tao Xie, and
  Ting Liu. 2020.
\newblock \href {https://doi.org/10.18653/v1/2020.emnlp-main.118} {Benchmarking
  meaning representations in neural semantic parsing}.
\newblock In \emph{{EMNLP}}.

\bibitem[{Guo et~al.(2019)Guo, Zhan, Gao, Xiao, Lou, Liu, and Zhang}]{Guo:2019}
Jiaqi Guo, Zecheng Zhan, Yan Gao, Yan Xiao, Jian{-}Guang Lou, Ting Liu, and
  Dongmei Zhang. 2019.
\newblock \href {https://www.aclweb.org/anthology/P19-1444/} {Towards complex
  text-to-sql in cross-domain database with intermediate representation}.
\newblock In \emph{Proceedings of the 57th Conference of the Association for
  Computational Linguistics, {ACL} 2019, Florence, Italy, July 28- August 2,
  2019, Volume 1: Long Papers}, pages 4524--4535.

\bibitem[{Guu et~al.(2017)Guu, Pasupat, Liu, and Liang}]{onlineMML/Guu:2017}
Kelvin Guu, Panupong Pasupat, Evan~Zheran Liu, and Percy Liang. 2017.
\newblock \href {https://doi.org/10.18653/v1/P17-1097} {From language to
  programs: Bridging reinforcement learning and maximum marginal likelihood}.
\newblock In \emph{Proceedings of the 55th Annual Meeting of the Association
  for Computational Linguistics, {ACL} 2017, Vancouver, Canada, July 30 -
  August 4, Volume 1: Long Papers}, pages 1051--1062. Association for
  Computational Linguistics.

\bibitem[{Hadsell et~al.(2006)Hadsell, Chopra, and LeCun}]{CL/Hadsell:2006}
Raia Hadsell, Sumit Chopra, and Yann LeCun. 2006.
\newblock \href {https://doi.org/10.1109/CVPR.2006.100} {Dimensionality
  reduction by learning an invariant mapping}.
\newblock In \emph{2006 {IEEE} Computer Society Conference on Computer Vision
  and Pattern Recognition {(CVPR} 2006), 17-22 June 2006, New York, NY, {USA}},
  pages 1735--1742. {IEEE} Computer Society.

\bibitem[{He et~al.(2016)He, Xia, Qin, Wang, Yu, Liu, and Ma}]{Xia:2016}
Di~He, Yingce Xia, Tao Qin, Liwei Wang, Nenghai Yu, Tie{-}Yan Liu, and
  Wei{-}Ying Ma. 2016.
\newblock \href
  {http://papers.nips.cc/paper/6469-dual-learning-for-machine-translation}
  {Dual learning for machine translation}.
\newblock In \emph{Advances in Neural Information Processing Systems 29: Annual
  Conference on Neural Information Processing Systems 2016, December 5-10,
  2016, Barcelona, Spain}, pages 820--828.

\bibitem[{Herzig and Berant(2019)}]{Herzig:2019}
Jonathan Herzig and Jonathan Berant. 2019.
\newblock \href {https://doi.org/10.18653/v1/D19-1394} {Don't paraphrase,
  detect! rapid and effective data collection for semantic parsing}.
\newblock In \emph{Proceedings of the 2019 Conference on Empirical Methods in
  Natural Language Processing and the 9th International Joint Conference on
  Natural Language Processing, {EMNLP-IJCNLP} 2019, Hong Kong, China, November
  3-7, 2019}, pages 3808--3818. Association for Computational Linguistics.

\bibitem[{Herzig et~al.(2021)Herzig, Shaw, Chang, Guu, Pasupat, and
  Zhang}]{CGIRSP/Herzig:2021}
Jonathan Herzig, Peter Shaw, Ming{-}Wei Chang, Kelvin Guu, Panupong Pasupat,
  and Yuan Zhang. 2021.
\newblock \href {http://arxiv.org/abs/2104.07478} {Unlocking compositional
  generalization in pre-trained models using intermediate representations}.
\newblock \emph{CoRR}, abs/2104.07478.

\bibitem[{Hochreiter and Schmidhuber(1997)}]{Hochreiter:1997}
Sepp Hochreiter and J{\"{u}}rgen Schmidhuber. 1997.
\newblock \href {https://doi.org/10.1162/neco.1997.9.8.1735} {Long short-term
  memory}.
\newblock \emph{Neural Computation}, 9(8):1735--1780.

\bibitem[{Huang et~al.(2021)Huang, Yang, Cao, Za{\"{\i}}ane, and
  Mou}]{Global/Huang:2021}
Chenyang Huang, Wei Yang, Yanshuai Cao, Osmar~R. Za{\"{\i}}ane, and Lili Mou.
  2021.
\newblock \href {http://arxiv.org/abs/2106.03376} {A globally normalized neural
  model for semantic parsing}.
\newblock \emph{CoRR}, abs/2106.03376.

\bibitem[{Iyyer et~al.(2017)Iyyer, Yih, and
  Chang}]{structuredLearning:/Iyyer:2017}
Mohit Iyyer, Wen{-}tau Yih, and Ming{-}Wei Chang. 2017.
\newblock \href {https://doi.org/10.18653/v1/P17-1167} {Search-based neural
  structured learning for sequential question answering}.
\newblock In \emph{Proceedings of the 55th Annual Meeting of the Association
  for Computational Linguistics, {ACL} 2017, Vancouver, Canada, July 30 -
  August 4, Volume 1: Long Papers}, pages 1821--1831. Association for
  Computational Linguistics.

\bibitem[{Jia and Liang(2016)}]{Robin:2016}
Robin Jia and Percy Liang. 2016.
\newblock \href {http://aclweb.org/anthology/P/P16/P16-1002.pdf} {Data
  recombination for neural semantic parsing}.
\newblock In \emph{Proceedings of the 54th Annual Meeting of the Association
  for Computational Linguistics, {ACL} 2016, August 7-12, 2016, Berlin,
  Germany, Volume 1: Long Papers}.

\bibitem[{Karpukhin et~al.(2020)Karpukhin, Oguz, Min, Lewis, Wu, Edunov, Chen,
  and Yih}]{CLQA/Karpukhin:2020}
Vladimir Karpukhin, Barlas Oguz, Sewon Min, Patrick S.~H. Lewis, Ledell Wu,
  Sergey Edunov, Danqi Chen, and Wen{-}tau Yih. 2020.
\newblock \href {https://doi.org/10.18653/v1/2020.emnlp-main.550} {Dense
  passage retrieval for open-domain question answering}.
\newblock In \emph{Proceedings of the 2020 Conference on Empirical Methods in
  Natural Language Processing, {EMNLP} 2020, Online, November 16-20, 2020},
  pages 6769--6781. Association for Computational Linguistics.

\bibitem[{Kate et~al.(2005)Kate, Wong, and Mooney}]{Kate:2005}
Rohit~J. Kate, Yuk~Wah Wong, and Raymond~J. Mooney. 2005.
\newblock \href {http://www.aaai.org/Library/AAAI/2005/aaai05-168.php}
  {Learning to transform natural to formal languages}.
\newblock In \emph{Proceedings, The Twentieth National Conference on Artificial
  Intelligence and the Seventeenth Innovative Applications of Artificial
  Intelligence Conference, July 9-13, 2005, Pittsburgh, Pennsylvania, {USA}},
  pages 1062--1068.

\bibitem[{Kim et~al.(2021)Kim, Yoo, and Lee}]{CL-BERT/Kim:2021}
Taeuk Kim, Kang~Min Yoo, and Sang{-}goo Lee. 2021.
\newblock \href {https://doi.org/10.18653/v1/2021.acl-long.197} {Self-guided
  contrastive learning for {BERT} sentence representations}.
\newblock In \emph{Proceedings of the 59th Annual Meeting of the Association
  for Computational Linguistics and the 11th International Joint Conference on
  Natural Language Processing, {ACL/IJCNLP} 2021, (Volume 1: Long Papers),
  Virtual Event, August 1-6, 2021}, pages 2528--2540. Association for
  Computational Linguistics.

\bibitem[{Kingma and Ba(2015)}]{Adam/Kingma:2015}
Diederik~P. Kingma and Jimmy Ba. 2015.
\newblock \href {http://arxiv.org/abs/1412.6980} {Adam: {A} method for
  stochastic optimization}.
\newblock In \emph{3rd International Conference on Learning Representations,
  {ICLR} 2015, San Diego, CA, USA, May 7-9, 2015, Conference Track
  Proceedings}.

\bibitem[{Lee et~al.(2021)Lee, Lee, and Hwang}]{CL4Gen/Lee:2021}
Seanie Lee, Dong~Bok Lee, and Sung~Ju Hwang. 2021.
\newblock \href {https://openreview.net/forum?id=Wga\_hrCa3P3} {Contrastive
  learning with adversarial perturbations for conditional text generation}.
\newblock In \emph{9th International Conference on Learning Representations,
  {ICLR} 2021, Virtual Event, Austria, May 3-7, 2021}. OpenReview.net.

\bibitem[{Li et~al.(2018)Li, Jiang, Shang, and Li}]{RL-para/Li:2018}
Zichao Li, Xin Jiang, Lifeng Shang, and Hang Li. 2018.
\newblock \href {https://doi.org/10.18653/v1/d18-1421} {Paraphrase generation
  with deep reinforcement learning}.
\newblock In \emph{Proceedings of the 2018 Conference on Empirical Methods in
  Natural Language Processing, Brussels, Belgium, October 31 - November 4,
  2018}, pages 3865--3878. Association for Computational Linguistics.

\bibitem[{Liu et~al.(2020)Liu, An, Lou, Chen, Lin, Gao, Zhou, Zheng, and
  Zhang}]{CompositionalGr/Liu:2020}
Qian Liu, Shengnan An, Jian{-}Guang Lou, Bei Chen, Zeqi Lin, Yan Gao, Bin Zhou,
  Nanning Zheng, and Dongmei Zhang. 2020.
\newblock \href
  {https://proceedings.neurips.cc/paper/2020/hash/83adc9225e4deb67d7ce42d58fe5157c-Abstract.html}
  {Compositional generalization by learning analytical expressions}.
\newblock In \emph{Advances in Neural Information Processing Systems 33: Annual
  Conference on Neural Information Processing Systems 2020, NeurIPS 2020,
  December 6-12, 2020, virtual}.

\bibitem[{Lu et~al.(2008)Lu, Ng, Lee, and Zettlemoyer}]{WeiLu:2008}
Wei Lu, Hwee~Tou Ng, Wee~Sun Lee, and Luke~S. Zettlemoyer. 2008.
\newblock \href {http://www.aclweb.org/anthology/D08-1082} {A generative model
  for parsing natural language to meaning representations}.
\newblock In \emph{2008 Conference on Empirical Methods in Natural Language
  Processing, {EMNLP} 2008, Proceedings of the Conference, 25-27 October 2008,
  Honolulu, Hawaii, USA, {A} meeting of SIGDAT, a Special Interest Group of the
  {ACL}}, pages 783--792.

\bibitem[{van~der Maaten and Hinton(2008)}]{tSNE2008}
Laurens van~der Maaten and Geoffrey Hinton. 2008.
\newblock Visualizing data using t-sne.
\newblock JMLR.

\bibitem[{Marzoev et~al.(2020)Marzoev, Madden, Kaashoek, Cafarella, and
  Andreas}]{unNLP/Marzoev:2020}
Alana Marzoev, Samuel Madden, M.~Frans Kaashoek, Michael~J. Cafarella, and
  Jacob Andreas. 2020.
\newblock \href {http://arxiv.org/abs/2004.13645} {Unnatural language
  processing: Bridging the gap between synthetic and natural language data}.
\newblock \emph{CoRR}, abs/2004.13645.

\bibitem[{van~den Oord et~al.(2018)van~den Oord, Li, and
  Vinyals}]{CPC/Oord:2018}
A{\"{a}}ron van~den Oord, Yazhe Li, and Oriol Vinyals. 2018.
\newblock \href {http://arxiv.org/abs/1807.03748} {Representation learning with
  contrastive predictive coding}.
\newblock \emph{CoRR}, abs/1807.03748.

\bibitem[{Oren et~al.(2020)Oren, Herzig, Gupta, Gardner, and
  Berant}]{CompositionalGr/Oren:2020}
Inbar Oren, Jonathan Herzig, Nitish Gupta, Matt Gardner, and Jonathan Berant.
  2020.
\newblock \href {https://doi.org/10.18653/v1/2020.findings-emnlp.225}
  {Improving compositional generalization in semantic parsing}.
\newblock In \emph{Findings of the Association for Computational Linguistics:
  {EMNLP} 2020, Online Event, 16-20 November 2020}, volume {EMNLP} 2020 of
  \emph{Findings of {ACL}}, pages 2482--2495. Association for Computational
  Linguistics.

\bibitem[{Qin et~al.(2021)Qin, Lin, Takanobu, Liu, Li, Ji, Huang, Sun, and
  Zhou}]{ERICA/Qin:2021}
Yujia Qin, Yankai Lin, Ryuichi Takanobu, Zhiyuan Liu, Peng Li, Heng Ji, Minlie
  Huang, Maosong Sun, and Jie Zhou. 2021.
\newblock \href {https://doi.org/10.18653/v1/2021.acl-long.260} {{ERICA:}
  improving entity and relation understanding for pre-trained language models
  via contrastive learning}.
\newblock In \emph{Proceedings of the 59th Annual Meeting of the Association
  for Computational Linguistics and the 11th International Joint Conference on
  Natural Language Processing, {ACL/IJCNLP} 2021, (Volume 1: Long Papers),
  Virtual Event, August 1-6, 2021}, pages 3350--3363. Association for
  Computational Linguistics.

\bibitem[{Qu et~al.(2021)Qu, Shen, Shen, Sajeev, Chen, and Han}]{CODA/Qu:2021}
Yanru Qu, Dinghan Shen, Yelong Shen, Sandra Sajeev, Weizhu Chen, and Jiawei
  Han. 2021.
\newblock \href {https://openreview.net/forum?id=Ozk9MrX1hvA} {Coda:
  Contrast-enhanced and diversity-promoting data augmentation for natural
  language understanding}.
\newblock In \emph{9th International Conference on Learning Representations,
  {ICLR} 2021, Virtual Event, Austria, May 3-7, 2021}. OpenReview.net.

\bibitem[{Rabinovich et~al.(2017)Rabinovich, Stern, and Klein}]{Rabin:2017}
Maxim Rabinovich, Mitchell Stern, and Dan Klein. 2017.
\newblock \href {https://doi.org/10.18653/v1/P17-1105} {Abstract syntax
  networks for code generation and semantic parsing}.
\newblock In \emph{Proceedings of the 55th Annual Meeting of the Association
  for Computational Linguistics, {ACL} 2017, Vancouver, Canada, July 30 -
  August 4, Volume 1: Long Papers}, pages 1139--1149.

\bibitem[{Shao et~al.(2020)Shao, Gong, Qi, Duan, and Lin}]{Shao:2020}
Bo~Shao, Yeyun Gong, Weizhen Qi, Nan Duan, and Xiaola Lin. 2020.
\newblock \href {https://doi.org/10.18653/v1/2020.coling-main.289} {Multi-level
  alignment pretraining for multi-lingual semantic parsing}.
\newblock In \emph{Proceedings of the 28th International Conference on
  Computational Linguistics, {COLING} 2020, Barcelona, Spain (Online), December
  8-13, 2020}, pages 3246--3256. International Committee on Computational
  Linguistics.

\bibitem[{Shen et~al.(2016)Shen, Cheng, He, He, Wu, Sun, and
  Liu}]{MinRisk/Shen:2016}
Shiqi Shen, Yong Cheng, Zhongjun He, Wei He, Hua Wu, Maosong Sun, and Yang Liu.
  2016.
\newblock \href {https://doi.org/10.18653/v1/p16-1159} {Minimum risk training
  for neural machine translation}.
\newblock In \emph{Proceedings of the 54th Annual Meeting of the Association
  for Computational Linguistics, {ACL} 2016, August 7-12, 2016, Berlin,
  Germany, Volume 1: Long Papers}. The Association for Computer Linguistics.

\bibitem[{Shin et~al.(2021)Shin, Lin, Thomson, Chen, Roy, Platanios, Pauls,
  Klein, Eisner, and Durme}]{FewShotSP/Shin:2021}
Richard Shin, Christopher~H. Lin, Sam Thomson, Charles Chen, Subhro Roy,
  Emmanouil~Antonios Platanios, Adam Pauls, Dan Klein, Jason Eisner, and
  Benjamin~Van Durme. 2021.
\newblock \href {http://arxiv.org/abs/2104.08768} {Constrained language models
  yield few-shot semantic parsers}.
\newblock \emph{CoRR}, abs/2104.08768.

\bibitem[{Su and Yan(2017)}]{Su:2017}
Yu~Su and Xifeng Yan. 2017.
\newblock \href {https://www.aclweb.org/anthology/D17-1127/} {Cross-domain
  semantic parsing via paraphrasing}.
\newblock In \emph{Proceedings of the 2017 Conference on Empirical Methods in
  Natural Language Processing, {EMNLP} 2017, Copenhagen, Denmark, September
  9-11, 2017}, pages 1235--1246.

\bibitem[{Wang et~al.(2021{\natexlab{a}})Wang, Ding, Li, and
  Zheng}]{CLINE/Wang:2021}
Dong Wang, Ning Ding, Piji Li, and Haitao Zheng. 2021{\natexlab{a}}.
\newblock \href {https://doi.org/10.18653/v1/2021.acl-long.181} {{CLINE:}
  contrastive learning with semantic negative examples for natural language
  understanding}.
\newblock In \emph{Proceedings of the 59th Annual Meeting of the Association
  for Computational Linguistics and the 11th International Joint Conference on
  Natural Language Processing, {ACL/IJCNLP} 2021, (Volume 1: Long Papers),
  Virtual Event, August 1-6, 2021}, pages 2332--2342. Association for
  Computational Linguistics.

\bibitem[{Wang et~al.(2015)Wang, Berant, and Liang}]{Wang:2015}
Yushi Wang, Jonathan Berant, and Percy Liang. 2015.
\newblock \href {https://www.aclweb.org/anthology/P15-1129/} {Building a
  semantic parser overnight}.
\newblock In \emph{Proceedings of the 53rd Annual Meeting of the Association
  for Computational Linguistics and the 7th International Joint Conference on
  Natural Language Processing of the Asian Federation of Natural Language
  Processing, {ACL} 2015, July 26-31, 2015, Beijing, China, Volume 1: Long
  Papers}, pages 1332--1342.

\bibitem[{Wang et~al.(2021{\natexlab{b}})Wang, Wang, Han, Lin, Hou, Liu, Li,
  Li, and Zhou}]{CLEVE/Wang:2021}
Ziqi Wang, Xiaozhi Wang, Xu~Han, Yankai Lin, Lei Hou, Zhiyuan Liu, Peng Li,
  Juanzi Li, and Jie Zhou. 2021{\natexlab{b}}.
\newblock \href {https://doi.org/10.18653/v1/2021.acl-long.491} {{CLEVE:}
  contrastive pre-training for event extraction}.
\newblock In \emph{Proceedings of the 59th Annual Meeting of the Association
  for Computational Linguistics and the 11th International Joint Conference on
  Natural Language Processing, {ACL/IJCNLP} 2021, (Volume 1: Long Papers),
  Virtual Event, August 1-6, 2021}, pages 6283--6297. Association for
  Computational Linguistics.

\bibitem[{Wong and Mooney(2007)}]{Wong:2007}
Yuk~Wah Wong and Raymond~J. Mooney. 2007.
\newblock \href {http://aclweb.org/anthology/P07-1121} {Learning synchronous
  grammars for semantic parsing with lambda calculus}.
\newblock In \emph{{ACL} 2007, Proceedings of the 45th Annual Meeting of the
  Association for Computational Linguistics, June 23-30, 2007, Prague, Czech
  Republic}.

\bibitem[{Wu et~al.(2021)Wu, Chen, Xin, Han, Sun, Zhang, Chen, Yang, and
  Cai}]{SSD/Wu:2021}
Shan Wu, Bo~Chen, Chunlei Xin, Xianpei Han, Le~Sun, Weipeng Zhang, Jiansong
  Chen, Fan Yang, and Xunliang Cai. 2021.
\newblock \href {https://doi.org/10.18653/v1/2021.acl-long.397} {From
  paraphrasing to semantic parsing: Unsupervised semantic parsing via
  synchronous semantic decoding}.
\newblock In \emph{Proceedings of the 59th Annual Meeting of the Association
  for Computational Linguistics and the 11th International Joint Conference on
  Natural Language Processing, {ACL/IJCNLP} 2021, (Volume 1: Long Papers),
  Virtual Event, August 1-6, 2021}, pages 5110--5121. Association for
  Computational Linguistics.

\bibitem[{Xiao et~al.(2016{\natexlab{a}})Xiao, Dymetman, and
  Gardent}]{DSP/Xiao:2016}
Chunyang Xiao, Marc Dymetman, and Claire Gardent. 2016{\natexlab{a}}.
\newblock \href {https://doi.org/10.18653/v1/p16-1127} {Sequence-based
  structured prediction for semantic parsing}.
\newblock In \emph{Proceedings of the 54th Annual Meeting of the Association
  for Computational Linguistics, {ACL} 2016, August 7-12, 2016, Berlin,
  Germany, Volume 1: Long Papers}. The Association for Computer Linguistics.

\bibitem[{Xiao et~al.(2016{\natexlab{b}})Xiao, Dymetman, and
  Gardent}]{structuredLearning:/Xiao:2016}
Chunyang Xiao, Marc Dymetman, and Claire Gardent. 2016{\natexlab{b}}.
\newblock \href {https://doi.org/10.18653/v1/p16-1127} {Sequence-based
  structured prediction for semantic parsing}.
\newblock In \emph{Proceedings of the 54th Annual Meeting of the Association
  for Computational Linguistics, {ACL} 2016, August 7-12, 2016, Berlin,
  Germany, Volume 1: Long Papers}. The Association for Computer Linguistics.

\bibitem[{Yih et~al.(2015)Yih, Chang, He, and
  Gao}]{structuredLearning:/Yih:2015}
Wen{-}tau Yih, Ming{-}Wei Chang, Xiaodong He, and Jianfeng Gao. 2015.
\newblock \href {https://doi.org/10.3115/v1/p15-1128} {Semantic parsing via
  staged query graph generation: Question answering with knowledge base}.
\newblock In \emph{Proceedings of the 53rd Annual Meeting of the Association
  for Computational Linguistics and the 7th International Joint Conference on
  Natural Language Processing of the Asian Federation of Natural Language
  Processing, {ACL} 2015, July 26-31, 2015, Beijing, China, Volume 1: Long
  Papers}, pages 1321--1331. The Association for Computer Linguistics.

\bibitem[{Yin et~al.(2018)Yin, Zhou, He, and Neubig}]{StructVAE/Yin:2018}
Pengcheng Yin, Chunting Zhou, Junxian He, and Graham Neubig. 2018.
\newblock \href {https://doi.org/10.18653/v1/P18-1070} {Structvae:
  Tree-structured latent variable models for semi-supervised semantic parsing}.
\newblock In \emph{Proceedings of the 56th Annual Meeting of the Association
  for Computational Linguistics, {ACL} 2018, Melbourne, Australia, July 15-20,
  2018, Volume 1: Long Papers}, pages 754--765.

\bibitem[{Zelle and Mooney(1996)}]{Zelle:1996}
John~M. Zelle and Raymond~J. Mooney. 1996.
\newblock \href {http://www.aaai.org/Library/AAAI/1996/aaai96-156.php}
  {Learning to parse database queries using inductive logic programming}.
\newblock In \emph{Proceedings of the Thirteenth National Conference on
  Artificial Intelligence and Eighth Innovative Applications of Artificial
  Intelligence Conference, {AAAI} 96, {IAAI} 96, Portland, Oregon, USA, August
  4-8, 1996, Volume 2.}, pages 1050--1055.

\bibitem[{Zhao et~al.(2020)Zhao, Sun, Cao, and Wan}]{Zhao:2020}
Yuanyuan Zhao, Weiwei Sun, Junjie Cao, and Xiaojun Wan. 2020.
\newblock \href {https://doi.org/10.18653/v1/2020.acl-main.606} {Semantic
  parsing for english as a second language}.
\newblock In \emph{Proceedings of the 58th Annual Meeting of the Association
  for Computational Linguistics, {ACL} 2020, Online, July 5-10, 2020}, pages
  6783--6794. Association for Computational Linguistics.

\end{thebibliography}
\bibliographystyle{acl_natbib}

\newpage

\usetikzlibrary{fit,positioning}
\usetikzlibrary{arrows}

\appendix
  \renewcommand{\appendixname}{Appendix~\Alph{section}}

\section{Appendix}

\begin{table}[htbp]
  \centering
  \resizebox{0.47\textwidth}{!}{
    \begin{tabular}{|l|c|c|}
    \hline
     & Overall test cases & Error cases \\
    \hline
    Jaccard index & 0.95 & 0.86 \\
    \hline
    Token coverage & 98.4\% & 90.1\% \\
    \hline
    Edit distance & 0.58 & 3.39\\
    \hline
    \end{tabular} }%
  \caption{Statistics of the distances between the predicted meaning representations and the correct ones.
  The error cases account for 13.6\% of the overall test dataset.
  }
  \label{ErrorStatistics}%
\end{table}%

\subsection{Error Case Statistics}
We use several indicators to measure the gap between the predicted meaning representation $y'$ and the correct mentioned representation $y$.
The basic \textsc{Seq2Seq} semantic parser is trained with MLE.
The statistical results of \textsc{Basketball} domain in  \textsc{Overnight} are shown in Table \ref{ErrorStatistics}.
The three statistical indicators are used to represent the similarity between $y'$ and $y$. 
The Jaccard index is calculated by:
\begin{equation}
J(y',y)=\frac{|y' \cap y|}{|y' \cup y|}
\end{equation}

We can see that the model can easily produce outputs very similar to the correct results.
Specifically, from 42.7\% of the error parses to the correct MRs  the edit distances  are only 1, and from 74.2\% of the error parses the edit distances of  are  $\le$ 3. 
This reveals that the predicted wrong MRs of the parsers are very close to the correct answers, but it is still hard for the parsers to differentiate the fine-grained semantics precisely.

\end{document}